\documentclass[letterpaper, 10 pt, conference]{ieeeconf}  % Comment this line out if you need a4paper

\IEEEoverridecommandlockouts                              % This command is only needed if 
\usepackage{graphicx}
\usepackage{cite}
\usepackage{picinpar}
\usepackage{amsmath, amssymb, mathrsfs}

\usepackage{url}
\usepackage{flushend}
\usepackage{colortbl}
\usepackage{soul}
\usepackage{multirow}
\usepackage{pifont}
\usepackage{color}
\usepackage{alltt}
\usepackage[hidelinks]{hyperref}
\usepackage{enumerate}
\usepackage{siunitx}
\usepackage{epstopdf}
\usepackage{pbox}
\usepackage{comment}

\usepackage{color}
\usepackage{xcolor}
\usepackage{amsmath,amssymb}
\usepackage{breqn}
\usepackage{mathtools}
\usepackage{layouts}
\usepackage{multicol}
\usepackage{booktabs}
\usepackage{multirow}
\usepackage{url}
\usepackage{threeparttable}
\usepackage[caption=false]{subfig}
\usepackage{siunitx}
\usepackage{bbding}
\usepackage{cases}
\usepackage{amsmath}

\usepackage{hyperref}
\hypersetup{
    colorlinks=true,
    linkcolor=blue, % 设置链接的颜色
    citecolor=blue, % 设置引用的颜色
    urlcolor=blue % 设置URL的颜色
}

\usepackage{CJKutf8}

\usepackage{algorithm,algpseudocode}
\title{\LARGE \bf
VIMS: A Visual-Inertial-Magnetic-Sonar SLAM System \\in Underwater Environments
}

\author{Bingbing Zhang$^{1,2,3}$,
    Huan Yin$^{3}$,
	Shuo Liu$^{1,*}$,
    Fumin Zhang$^{3}$,
	and Wen Xu$^{1,4}$% <-this % stops a space
\thanks{This work was supported by the National Natural Science Foundation Major Research Instrumentation Project titled ``Intelligent Agile Ocean Stereoscopic Observation Instrument'' (42227901),  the ``Pioneer'' and ``Leading Goose'' R\&D Program of Zhejiang (2022C03041 and 2023C03124), and the Strategic Priority Research Program of the Chinese Academy of Sciences under the grant number XDA22040202.}% <-this % stops a space
\thanks{$^{1}$Key Laboratory of Ocean Observation-Imaging Testbed of Zhejiang Province, Zhejiang University, Zhoushan, 316021, China        {\tt\small \{zhangbb, shuoliu, wxu\}@zju.edu.cn}}%
\thanks{$^{2}$Interdisciplinary Student Training Platform for Marine areas, Zhejiang University, Hangzhou, 310027, China}%
\thanks{$^{3}$Department of Electronic and Computer Engineering, The Hong Kong University of Science and Technology, Hong Kong SAR {\tt\small \{eehyin, eefumin\}@ust.hk}}
\thanks{$^{4}$The Institute of Deep-Sea Science and Engineering, Chinese Academy of Sciences, Sanya, 572000, Hainan, China}%
\thanks{*Corresponding author}
}

\begin{document}
		\begin{figure*}[!b]  % The placement specifier [!b] positions the box at the bottom of the page
		\centering
		\fbox{
			\parbox{\textwidth}{
				\small  % Smaller font size
				\textbf{Copyright Notice} \\
				This work has been accepted for publication at the IEEE/RSJ International Conference on Intelligent Robots and Systems (IROS 2025). Copyright may be transferred to IEEE upon final publication. This is the author's version of the work. The definitive version will be available in the official IEEE proceedings. Personal use is permitted, but permission from IEEE is required for all other uses, including reprinting/republishing.
			}
		}
	\end{figure*}
\maketitle
\thispagestyle{empty}
\pagestyle{empty}

%%%%%%%%%%%%%%%%%%%%%%%%%%%%%%%%%%%%%%%%%%%%%%%%%%%%%%%%%%%%%%%%%%%%%%%%%%%%%%%%
% Firstly, underwater vehicles experience minimal acceleration due to higher drag, compromising scale estimations. Secondly, visual place recognition is hindered by unstructured environments, limited visual range, and constrained computational resources onboard AUVs. Thirdly, VINS involves a challenging trade-off between image matching robustness during loop closure verification after place recognition and feature-tracking performance. 

\begin{abstract}
In this study, we present a novel simultaneous localization and mapping (SLAM) system, VIMS, designed for underwater navigation. Conventional visual-inertial state estimators encounter significant practical challenges in perceptually degraded underwater environments, particularly in scale estimation and loop closing. To address these issues, we first propose leveraging a low-cost single-beam sonar to improve scale estimation. Then, VIMS integrates a high-sampling-rate magnetometer for place recognition by utilizing magnetic signatures generated by an economical magnetic field coil. Building on this, a hierarchical scheme is developed for visual-magnetic place recognition, enabling robust loop closure. Furthermore, VIMS achieves a balance between local feature tracking and descriptor-based loop closing, avoiding additional computational burden on the front end. Experimental results highlight the efficacy of the proposed VIMS, demonstrating significant improvements in both the robustness and accuracy of state estimation within underwater environments.
\end{abstract}
% in orb slam and vins-mono frameworks, 

\section{Introduction}

% \begin{figure}[thpb]
% 	\centering
% 	%	\setlength{\abovecaptionskip}{-20pt}
% 	%	\setlength{\belowcaptionskip}{50pt}
% 	\includegraphics[width=\linewidth]{./figs/empty2.png}
% 	\caption{Affiliations}
% 	\label{figslam}
% \end{figure}
\begin{figure}
    \centering
    \includegraphics[width=0.99\linewidth]{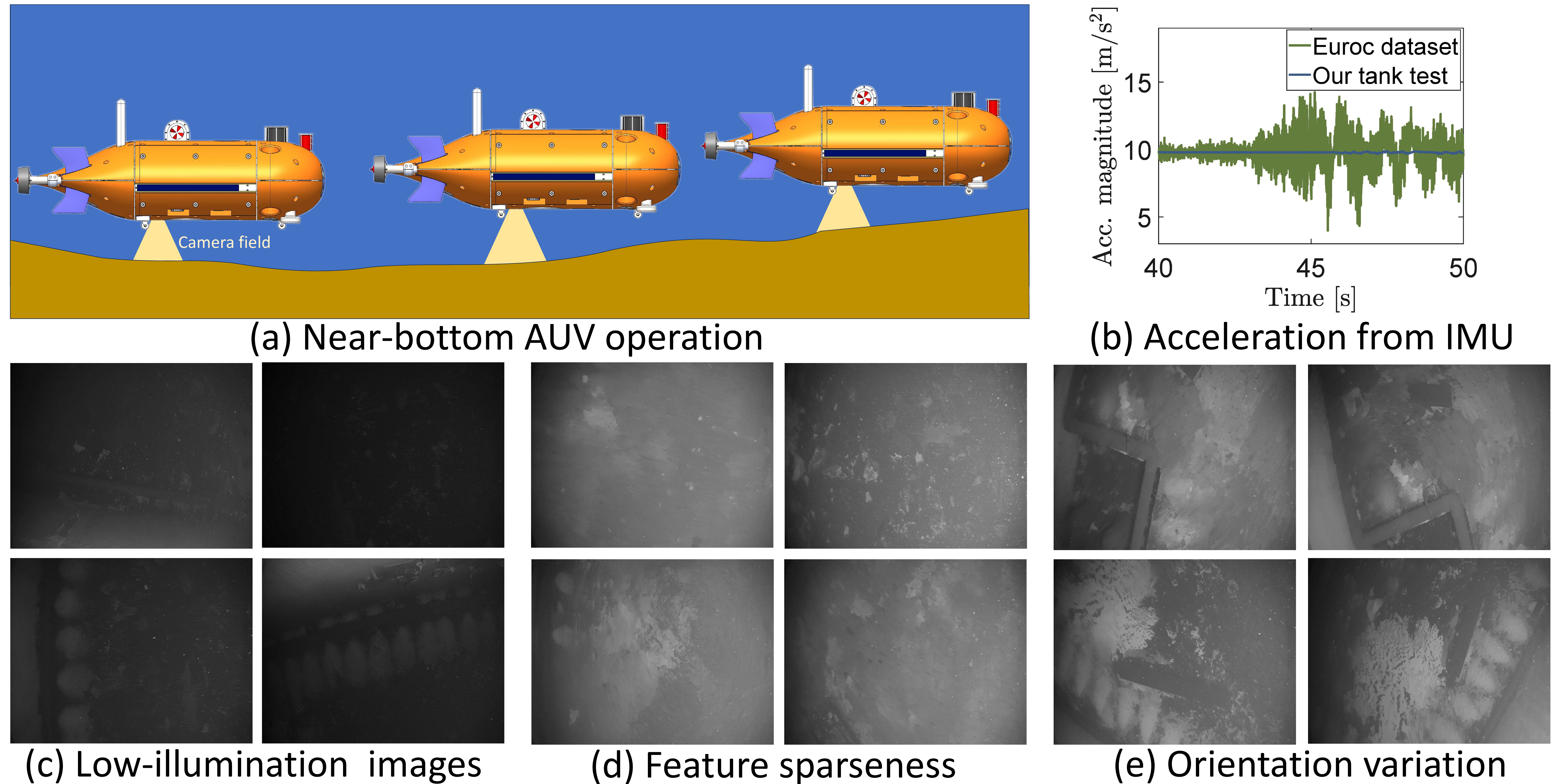}
    \caption{Challenges of visual-inertial navigation in underwater environments.
(a) A graphical illustration of AUV operation near the seafloor.
(b) The self-acceleration in AUV motion is significantly smaller than that observed in conventional scenarios, such as the Euroc dataset~\cite{Burri25012016}, as shown in the figure. This results in inaccurate motion scale estimation for the AUV.
(c) Low-illumination images arise when the AUV is positioned far from the seafloor.
(d) Underwater environments often suffer from sparse features.
(e) Loop closing is significantly affected by orientation variation when the AUV revisits nearby locations.}
    \label{fig_teaser}
\end{figure}

% Acceleration magnitudes are obtained from the IMU, encompassing both gravity and self-acceleration effects. The self-acceleration magnitude is determined by analyzing deviations from the expected gravitational strength of approximately 9.8 m/s².

\begin{figure*}[t]
\centering
\includegraphics[width=0.95\linewidth]{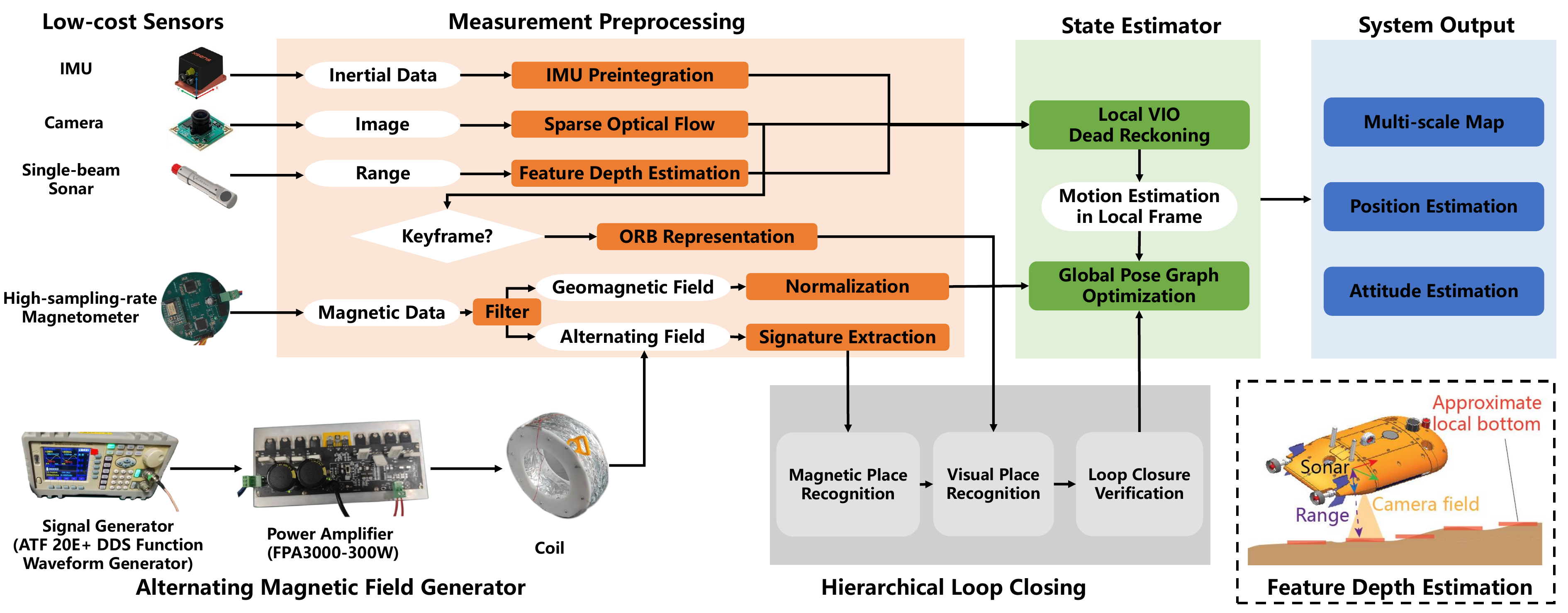}
\caption{System overview. The VIMS begins with measurement processing (Section \ref{sectMeasurement}), followed by dead reckoning and global optimization for state estimation (Section \ref{sectState}). In the mapping part, the system incrementally constructs and continuously updates a multi-scale magnetism and visual map (Section \ref{sectMulti}). The feature depth estimation module using sonar is illustrated  graphically at the bottom right.}
\label{figslam}
\end{figure*}
% \yh{could be better, I suggest you upload the ppt file on overleaf. I will revise it in my free time.}  \bb{Now I use ppt file in the ppt folder.}

\label{sectIntro}

\subsection{Motivations}
Autonomous underwater vehicles (AUVs) rely on precise positioning to ensure safe and efficient task execution~\cite{9783050,xu2023robust,zhang2023autonomous}. The lack of GPS signals in underwater environments presents significant challenges for vehicular localization. Existing methods, such as acoustic localization~\cite{9783050} and Doppler velocity log (DVL)-based dead reckoning~\cite{paterson2024nonlinear}, are often limited by factors such as weight and cost, thus restricting their widespread adoption on AUVs~\cite{xu2020integrated}. A promising alternative is the use of visual-inertial navigation systems (VINS), which have proven effective on self-driving cars and drones in GPS-denied environments~\cite{qin2018vins}. 

In classical VINS, visual-inertial odometry (VIO)~\cite{4209642,forster2014svo,forster2016svo,xu2020integrated} continuously estimates vehicle poses, while the loop closing module reduces uncertainty when revisiting known locations~\cite{leutenegger2015keyframe,campos2021orb}. However, deploying conventional VINS in underwater environments introduces unique and practical challenges, as illustrated in Figure~\ref{fig_teaser}. These challenges are summarized as follows:

\begin{enumerate}

\item \textbf{Inaccurate scale estimation.} Acceleration measurements play a crucial role in scale estimation in VINS. Accurate motion scale estimation requires sufficient self-acceleration excitation to mitigate the effects of noise and bias. However, AUVs operate in a denser medium with increased drag and are further constrained by lower power-to-weight ratios~\cite{9525214,huang2024self}. These factors restrict AUV motion, thereby reducing the contribution of acceleration measurements to scale estimation, which can potentially result in estimation failures in certain cases~\cite{von2022dm}.

\item \textbf{Poor place recognition performance.} The loop closing performance is influenced by multiple factors, including a lack of distinguishable local features~\cite{10323517}, low-illumination images, and orientation variations~\cite{wang2023real,song2024turtlmap}. Additionally, constrained computational resources exacerbate the risk of place recognition aliasing.

\item \textbf{Balancing feature tracking and loop closing.}  For local feature tracking, sparse optical flow-based methods~\cite{qin2018vins, xu2020integrated} generally outperform descriptor-based image matching~\cite{campos2021orb} in low-texture environments~\cite{singh2024opti,9779491}. However, using simple features will make loop closing highly sensitive to orientation changes, especially for downward-facing cameras commonly used in underwater navigation.  Seamlessly  integrating robust local feature tracking with reliable feature description for underwater SLAM remains a challenge.
\end{enumerate}

% This challenge complicates the development of SLAM systems that balance effective feature tracking with robust loop closing in underwater environments. Additionally, deploying such a visual system on resource-constrained AUVs presents further challenges.

\subsection{Contributions}

Given the challenges above, we design and construct a novel underwater SLAM system for AUVs named VIMS, which integrates magnetic, visual, inertial, and single-beam sonar measurements. 

To address the challenge of scale estimation, we deploy a \textit{low-cost single-beam sonar} to constrain feature depth. Existing works have utilized DVLs~\cite{kim2013real,xu2021underwater,9561537,huang2023tightly,huang2024self}, scanning profiling sonar~\cite{rahman2022svin2}, or imaging sonar~\cite{jang2021multi,bucci2022evaluation,yang2024acoustic,singh2024opti,zhang2024integration} to aid vision-based perception and estimation. However, these sensors are expensive for commonly used AUVs. Alternatively, some studies have employed low-cost pressure sensors for scale estimation~\cite{ding2023rd,9525214,9782141,ding2024underwater}, but these are impractical in scenarios with small vertical motion. In this study, we utilize a single-beam sonar to directly obtain feature depth measurements, under the assumption of a locally flat bottom. Existing works in~\cite{xu2020integrated, roznere2020underwater} also use single-beam sonar to aid scale estimation. Our method differs from~\cite{xu2020integrated} by eliminating the need for a high-accuracy gyroscope and a sliding-window strategy for state estimation. Unlike~\cite{roznere2020underwater}, we leverage the local flatness of the underwater bottom and incorporate additional sensors to ensure global alignment with the world coordinate system.

% , reducing noise and outliers for improved accuracy and robustness in state estimation

For place recognition, our previous work~\cite{10778419} demonstrates the effectiveness of integrating vision with alternating magnetic fields using off-the-shelf magnetic angular rate and gravity (MARG) sensors. This success inspires further investigation into visual-magnetic loop closing for more realistic and challenging scenarios. To reduce measurement sampling time, this study equips the vehicle with a \textit{high-sampling-rate magnetometer}, enabling hierarchical place recognition integrated with visual imagery, even during high-speed movement and operation in more demanding environments. An extremely-low-frequency (ELF) magnetic field is generated by a coil situated within the operation area, as illustrated in Figure \ref{figslam}. This field creates distinct 3D alternating magnetic field intensity vectors at different locations, enriching navigation cues and effectively aiding visual place recognition.

To enhance performance in low-texture environments, we apply optical flow to track Shi-Tomasi corners~\cite{shi1994good} in real-time. For loop closing, we employ ORB descriptors\cite{rublee2011orb} to represent places, performing descriptor extraction only during the place recognition stage. Existing studies recommend the use of GPUs for real-time complex feature description\cite{9779491}, but most underwater vehicles lack GPU-enabled processors~\cite{monterroso2023autonomous}. In this study, the implemented VIMS relies solely on the CPU, ensuring broader accessibility and compatibility across a wide range of underwater platforms.

In summary, the key contributions are three-fold: 
\begin{itemize}
\item Leveraging single-beam sonar altitude measurements to improve scale estimation and providing a low-cost dead reckoning solution with acceptable accuracy.
\item Introducing high-sampling-rate magnetometer measurements, as an extension of~\cite{10778419}. We design a hierarchical place recognition that integrates magnetic and visual signatures.
\item Enhancing VINS with ORB descriptors to address orientation variations without adding computational burden to the real-time front-end.
\end{itemize}

%\begin{figure}[t]
%	\centering
%	\includegraphics[width=\linewidth]{./figs/01introduction/generator.pdf}
%	\caption{Alternating magnetic field generator.}
%	\label{figGen}
%\end{figure}

\section{Notation and Assumptions}
\label{sec:notation}

We first define the following notations and frames. The superscript $(\cdot)^w$ represents the world frame (ENU convention), $(\cdot)^b$ for the body frame (aligned with the IMU), and $(\cdot)^c$ for the camera frame. Rotations are represented by matrices, $\mathbf{R}$, or Hamilton quaternions, $\mathbf{q}$. For example, $\mathbf{R}_b^w$ and $\mathbf{p}_b^w$ are the rotation and translation from the body to the world frame. The corresponding transformation matrix is defined as $
\mathbf{T}_b^w = \left[
\begin{matrix}
\mathbf{R}_b^w&\mathbf{p}_b^w\\
\mathbf{0}&1
\end{matrix}
\right]
$. The notation $(\cdot)[k]$ refers to the state or measurement at the moment of capturing the $k$-th image, while $(\cdot)(t)$ refers to the state or measurement at time $t$. The gravity vector in the world frame is $\mathbf{g}^w=[0,0,-g]^\top$. Noisy measurements are denoted by $\tilde{(\cdot)}$ and estimated quantity by $\hat{(\cdot)}$.

We assume that the AUV operates above a locally flat bottom, which is a common scenario in underwater environments~\cite{xu2020integrated}. Additionally, the vehicle is equipped with a pre-calibrated, offset-free magnetometer. The extrinsic parameters of the IMU-camera $\left(\mathbf{R}_b^c, \mathbf{p}_b^c\right)$ and IMU-sonar $\left(\mathbf{R}_b^s, \mathbf{p}_b^s\right)$ are assumed to be pre-calibrated and known.

%In the remainder of this paper, the map is composed of multiple submaps, each associated with a sensing modality and given by 
%\begin{equation}
%	submap^k(t) = \left[\hat{\mathbf{p}}_{b\left(t\right)}^l, \hat{\mathbf{q}}_{b\left(t\right)}^l, \hat{\mathbf{p}}_{b\left(t\right)}^w,
%	\hat{\mathbf{q}}_{b\left(t\right)}^w,
%	 {sig}^k\left(\mathbf{p}_{b\left(t\right)}^l\right)\right]
%\end{equation}
%where $\left[\hat{\mathbf{p}}_{b\left(t\right)}^l,  \hat{\mathbf{q}}_{b\left(t\right)}^l\right]$ is the pose estimate at the time $t$ in the local frame from dead reckoning, while $\left[\hat{\mathbf{p}}_{b\left(t\right)}^w, \hat{\mathbf{q}}_{b\left(t\right)}^w\right]$ is with respect to the world frame obtained by global pose graph optimization.  ${sig}^k\left(\mathbf{p}_{b\left(t\right)}^l\right)$ 
%is the signature encoding the local position  ${\mathbf{p}}_{b\left(t\right)}^l$.

\section{Measurement Preprocessing}
\label{sectMeasurement}

We introduce the preprocessing procedures in this section. We employ the same front-end modules from VINS-Mono~\cite{qin2018vins} for visual processing and IMU preintegration. Specifically, optical flow is used to track Shi-Tomasi corners across consecutive frames, establishing inter-frame data associations, while inertial measurements are pre-integrated between image frames. For range measurements from the sonar, a median filter is applied to reject outliers caused by multi-path effects. 

In contrast to our previous work~\cite{10778419}, this study employs a self-developed high-sampling-rate magnetometer. It reduces aliasing effects caused by high-frequency noise~\cite{Signals} and enhances signal processing precision during short sampling periods in high-speed motion. The hardware details are provided in Section \ref{sectSetup}. The data processing procedure follows the same methodology as in~\cite{10778419}, with a brief overview provided below.

The real-time magnetic measurements are the sum of the geomagnetic field and the alternating field, as expressed in the equation:
\begin{equation} \tilde{\mathbf{B}} = \mathbf{B}_e^b + \mathbf{R}_w^b \mathbf{h}(\mathbf{p}_b^w) + \mathbf{n}_{\mathbf{B}}
\end{equation}
in which $\mathbf{B}_e^b = \mathbf{R}_w^b \mathbf{B}_e^w$ represents the geomagnetic field in the body frame, $\mathbf{h}(\mathbf{p}_b^w)$ is a function relating the alternating field to the position $\mathbf{p}_b^w$, and $\mathbf{n}_{\mathbf{B}}$ denotes Gaussian white noise. Magnetic measurements are processed using a simple moving average filter to separate the geomagnetic data $\mathbf{R}_w^b \mathbf{B}_e^w$ and the alternating magnetic field $\mathbf{R}_w^b \mathbf{h}(\mathbf{p}_b^w)$ in $\tilde{\mathbf{B}}$. The alternating magnetic field amplitude vector, ${sig}^m (\mathbf{p}_b^w)= [B_E^{alt}, B_N^{alt}, B_U^{alt}]^\top $, with E-N-U components is then derived by applying FFT to $\mathbf{h}(\mathbf{p}_b^w)$. In this way, we obtain the magnetic signature ${sig}^m (\mathbf{p}_b^w)$. For further details, refer to our previous work~\cite{10778419}.

\section{State Estimation in VIMS} 
\label{sectState}

With the preprocessed measurements, the state estimation framework consists of a local VIO for motion estimation and a global pose graph for pose optimization.

\subsection{Motion Estimation by Local VIO }
\label{sectVio}

Our approach enhances the sliding window-based VIO module of VINS-Mono~\cite{qin2018vins} by integrating altitude data from a single-beam sonar. The VIO module estimates the 6-DoF vehicle state within sliding windows, given by:
\begin{equation}
\begin{aligned}
    {\mathcal{X}^l}&=[\mathbf{x}^l [0],\mathbf{x}^l [1],\ldots, \mathbf{x}^l [n], \lambda_0, \lambda_1,\ldots, \lambda_m]\\
    \mathbf{x}^l [k]&=[\mathbf{p}_{b[k]}^l,\mathbf{v}_{b[k]}^l,\mathbf{q}_{b[k]}^l,\mathbf{b}_a,\mathbf{b}_g],k\in [0,n]
\end{aligned}
\end{equation}
where $\mathbf{x}^l [k]$ denotes the IMU state at the capture of the $k$-th image, including position, velocity, orientation in the local reference frame $l$, as well as accelerometer  bias and gyroscope bias. $n$ is the window length, $m$ the number of tracked features. 

It is worth noting that $ \lambda_j $ represents the distance of the $ j $-th feature from its first observation, which differs from most VIO systems that estimate inverse depth~\cite{civera2008inverse} (e.g.,~\cite{qin2018vins}). In traditional VIO systems, acceleration measurements contribute to motion scale estimation between two frames. Based on this, absolute feature depth can be determined by the parallax between consecutive camera frames. The parallax exhibits good linearity with inverse depth, especially for features at large distances~\cite{civera2008inverse}. However, for AUVs operating in underwater environments, features at greater distances become invisible due to light attenuation, so we only consider features within a limited range of distances. More importantly, minimal self-acceleration diminishes the effectiveness of acceleration measurements in scale estimation. Therefore, in the proposed VIMS, single-beam sonar plays a dominant role in scale estimation. Since range measurements are more linearly related to feature depth, we adopt direct depth parametrization  instead of inverse depth, following the model used in most ranging sensors~\cite{10833750,s18051386}.

The position of the $j$-th feature in the local frame, first observed by the $k$-th image, is given by:
\begin{equation}
{\mathbf{f}}^l(\lambda_j)={\mathbf{T}}_{b[k]}^{l}\mathbf{T}_{c}^{b} {\mathbf{f}}^c(\lambda_j)
\end{equation}
where ${\mathbf{f}}^c(\lambda_j)$ represents the expected feature point position in the camera frame given feature depth $\lambda_j$, and ${\mathbf{f}}^l(\lambda_j)$ denotes its position in the local frame. Sonar range data measures the distance $d[k]$ from the sonar to the bottom along the signal path. The signal path's intersection with the seabed is given by
\begin{equation}
\tilde{\mathbf{d}}^l[k]=\mathbf{T}_{b[k]}^{l}\mathbf{T}_{s}^{b} 
\left[\begin{matrix}
    0&
    0&
    \tilde{d}[k]
\end{matrix}\right]^\top 
% \text{.}
\end{equation}
\par Assuming the local bottom planarity, as stated in Section~\ref{sec:notation}, these altitude measurements add constraints for positioning each feature. The residual is formulated as $
(\tilde{\mathbf{d}}^l[k])_3 - ({\mathbf{f}}^l(\lambda_j))_3
$, where $(\cdot)_3$ is the third element of the vector.

\begin{figure}[t]
\centering
\includegraphics[width=\linewidth]{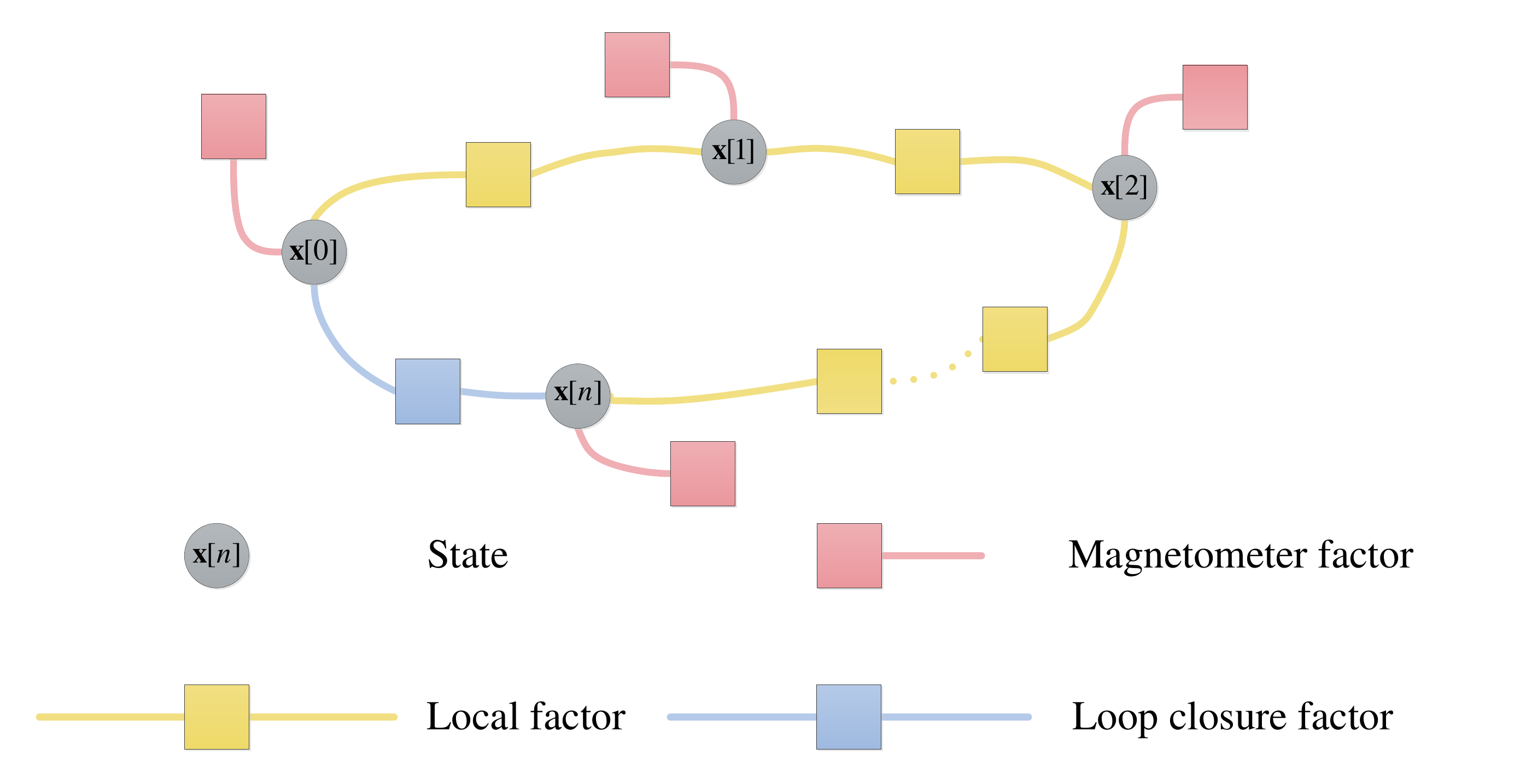}
\caption{Graphical illustration of global pose graph optimization. Each node represents a keyframe state in the world frame, including position and orientation. Edges connecting two consecutive nodes are local constraints derived from VIO. Other edges represent global constraints from magnetic measurements and loop closures.}
\label{figGlobal}
\end{figure}

\subsection{Global Pose Graph Optimization}

Pose graph optimization estimates global poses of all keyframes as $\mathcal{X}=\left\{\mathbf{x}^w [0], \mathbf{x}^w [1], \ldots, \mathbf{x}^w [n]\right\}$, where $\mathbf{x}^w [i]$ includes position and orientation in the world frame. The objective is to maximize the measurement probabilities with the assumption of Gaussian distributions on measurements. This results in the optimization problem as follows: 
\begin{equation}
\hat{\mathcal{X}} ={\arg \min\limits_{\mathcal{X}} } \sum_{i=0}^n \sum_{k \in \mathbf{S}}\left\|\tilde{\mathbf{z}}^k [i]-\mathbf{g}_i^k(\mathcal{X})\right\|_{\mathbf{\Omega}^k [i]}^2
\end{equation}
where $\mathcal{S}$ is the set of measurements, including local VIO estimates, magnetometer measurements, and loop closure results. $\mathbf{g}_i^k(\mathcal{X})$ is the expected $i$-th measurement of type $k$. 

Figure \ref{figGlobal} illustrates a factor graph representing the global pose graph optimization in VIMS. Magnetometer factors align local poses with global coordinates using Earth's magnetic field. Loop closure factors correct poses by detecting revisited locations, ensuring global consistency throughout the trajectory. In the following section, we will detail the mapping framework and our proposed loop closing scheme in underwater environments.

\section{Multi-Scale Mapping and Loop Closing}
\label{sectMulti}
Loop closure factors are crucial for maintaining the accuracy  of the pose graph, and ensuring consistent mapping~\cite{yin2024survey}. VIMS employs a multi-modal multi-scale map and hierarchical place recognition strategy to enhance loop closing performance. The key improvements compared to our previous work~\cite{10778419} include the integration of ORB descriptors~\cite{rublee2011orb} and a reduction in the magnetic measurement sampling time. Figure \ref{figmultiscale} presents an overview of the proposed mapping framework and loop closing.
%\yh{what is counterparts? better use other common words}

% We incorporate ORB features into the VINS framework~\cite{qin2018vins} and present a complete and multi-modal mapping framework.

\subsection{Multi-Scale Map Construction}

The pose graph associates historical poses with their corresponding local submaps, which include both visual and magnetic submaps. Due to differences in frequency and spatial variation, magnetic submaps are larger than visual submaps, enabling each magnetic submap to associate with multiple visual submaps.

\subsubsection{Visual Submaps}
\label{secVisual}
Visual submaps are added to the pose graph using selected keyframes. We use ORB descriptors to encode salient corners at different scales of the image pyramid, which could enhance feature representation and reduce orientation effects on loop closing (Figure \ref{figORB}). Each visual submap, denoted as the $i$-th submap, is characterized by:
\begin{equation}
\begin{aligned}
    submap^v [i]&=\left[\hat{\mathbf{p}}_{b[i]}^l,\hat{\mathbf{q}}_{b[i]}^l,\hat{\mathbf{p}}_{b[i]}^w,\hat{\mathbf{q}}_{b[i]}^w, {sig}^v\left(\mathbf{p}_{b\left[i\right]}^l\right)\right]\\
    {sig}^v\left(\mathbf{p}_{b\left[i\right]}^l\right)&=\mathcal{D}_i(u,v,des)
\end{aligned}
\end{equation} 
where $\mathcal{D}_i(u,v,des)$ represents the set of features from the image associated with the $i$-th visual submap. Each element contains 2-D pixel coordinates $(u,v)$ and an ORB descriptor, denoted as $des$. 

\begin{figure}[t]
\centering
\includegraphics[width=\linewidth]{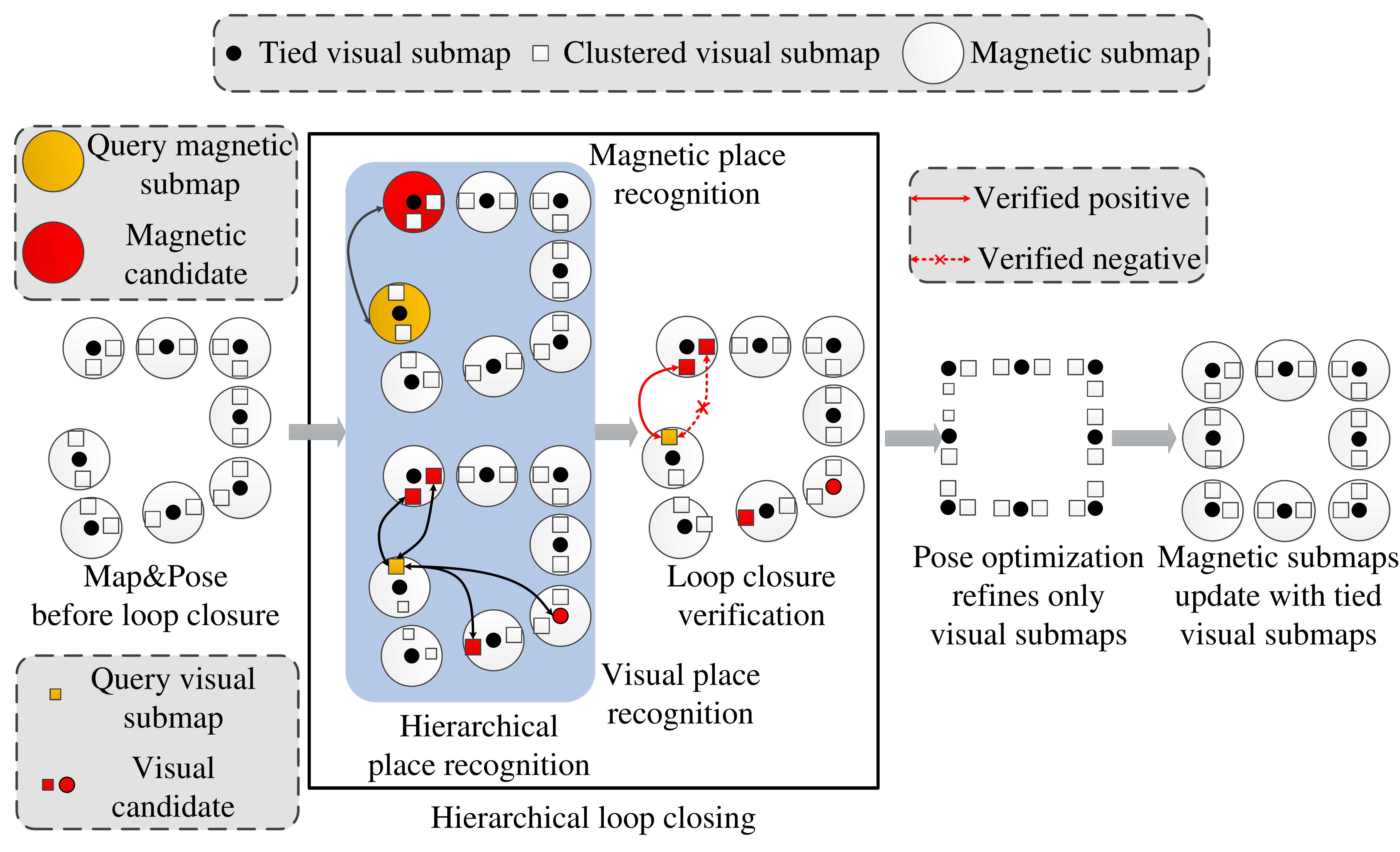}
\caption{Multi-scale mapping and hierarchical loop closing in VIMS. A magnetic submap is linked to multiple visual submaps, which include a tied visual submap and clustered visual submaps. If a submap passes magnetic place recognition, all clustered visual submaps are considered candidates for visual place recognition. A loop is deemed valid only after successfully passing magnetic place recognition, visual place recognition, and loop closure verification. After pose graph optimization, the pose associated with the magnetic submap is updated to match the pose of the tied visual submap.
}
\label{figmultiscale}
\end{figure}

\subsubsection{Magnetic Submaps}
\label{secMagMap}

Generally, the moving average filter and FFT require a specified collection period to reduce noise during the extraction of magnetic signatures. A longer collection time may, however, decrease the frequency of signature production. Using our self-designed high-sampling-rate magnetometer, as depicted in Figure \ref{figMagnetometer}, a 2-second collection period yields good results based on extensive testing. This approach represents an advancement over our previous work~\cite{10778419}, where MARG sensors with a 50-100 Hz sampling rate necessitated a collection period of 5-10 seconds to achieve satisfactory signal processing outcomes. Furthermore, we employ a sliding-window technique with overlapping windows, achieving a 0.4-second output interval. Integrating a high-sampling-rate magnetometer significantly supports higher operational speeds for the AUV.

Magnetic signatures are tied to the temporally nearest visual submap, ensuring that the positions of magnetic submaps are updated after global optimization, as shown in Figure \ref{figmultiscale}. We partition signatures into distinct submaps and utilize a sparse representation, encoding each submap with the smoothed value of the signature at the median point of the corresponding signatures (as detailed in our work~\cite{10778419}). This could enhance place recognition efficiency and reduce map consumption. We refine the magnetic signature ${sig}^m\left(\mathbf{p}_{b\left(t_k\right)}^l\right)$ to ${sig}^{m*}\left(\mathbf{p}_{b\left(t_k\right)}^l\right)$, adding a radius term that specifies the submap boundaries within which the maximum variation remains below a predefined threshold. This enhanced magnetic submap is then linked to the $i$-th visual submap and is given by
\begin{equation}
\label{eqSubmap}
submap^m [i]= \left[\hat{\mathbf{p}}_{b[i]}^l, \hat{\mathbf{p}}_{b[i]}^w, {sig}^{m*}\left(\mathbf{p}_{b[i]}^l\right)\right]\text{.}
\end{equation}
%subsubsection{Clustering Visual Submaps into Magnetic Submaps} 
\par The radius assigned to each magnetic submap guides the clustering of nearby visual submaps, creating a multi-scale map. An example is presented in Figure \ref{figTankLandmark}. 
%\yh{this subsubsection is too short. Please merge it into previous ones.}

\subsection{Hierarchical Place Recognition}

VIMS employs a hierarchical place recognition mechanism that integrates both magnetic and visual cues to guarantee the reliability:
\begin{itemize}

\item \textbf{Magnetic Place Recognition:} We compare current magnetic signature to existing submaps using the Euclidean norm, ${\|{sig}^{m}\left(\mathbf{p}_{b[i]}^l\right)-{sig}^{m}\left(\mathbf{p}_{b[j]}^l\right)\|_2}$, to measure signature differences. VIMS identifies potential matches by selecting magnetic submaps with the smallest signature differences.

\item \textbf{Visual Place Recognition:} We utilize classical DBoW2~\cite{galvez2012bags}, which uses word-like descriptors to search the visual database efficiently. DBoW2 identifies loop closure candidates from visual submaps, focusing on regions highlighted by magnetic candidates to ensure precise candidate selection.

\end{itemize}

\begin{figure}[t]
    \centering
    \includegraphics[width=0.75\linewidth]{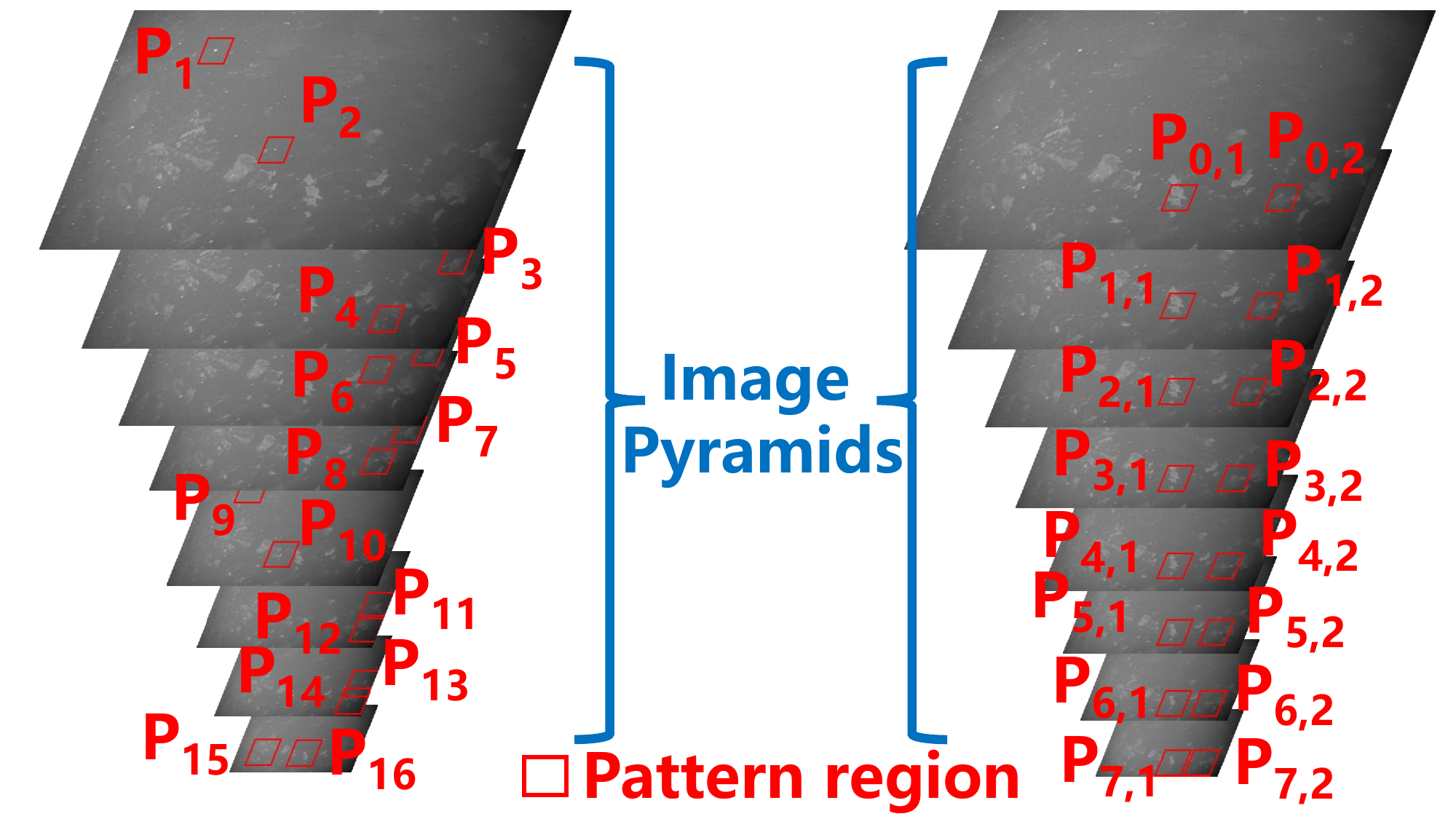}   
    
    \caption{Feature detection strategy. \textbf{Left:} Conventional corner selection. \textbf{Right:} Corner replication applied in VIMS. In the conventional ORB detector~\cite{rublee2011orb}, the most salient corners at each scale are selected. The replication method, on the other hand, replicates certain corners across scales to address scale variations. For visual submap representation (Section \ref{secVisual}), the conventional corner selection strategy is used. Meanwhile, in the sliding window feature description (Section \ref{secLoop}), the replication method is applied, as only the 3D locations of tracked points are estimated.}
    \label{figORB}
\end{figure}

\begin{figure}[t]
\centering
\subfloat[Haihong \uppercase\expandafter{\romannumeral2} AUV and tank]{\label{figPlatform}\includegraphics[width=0.45\linewidth]{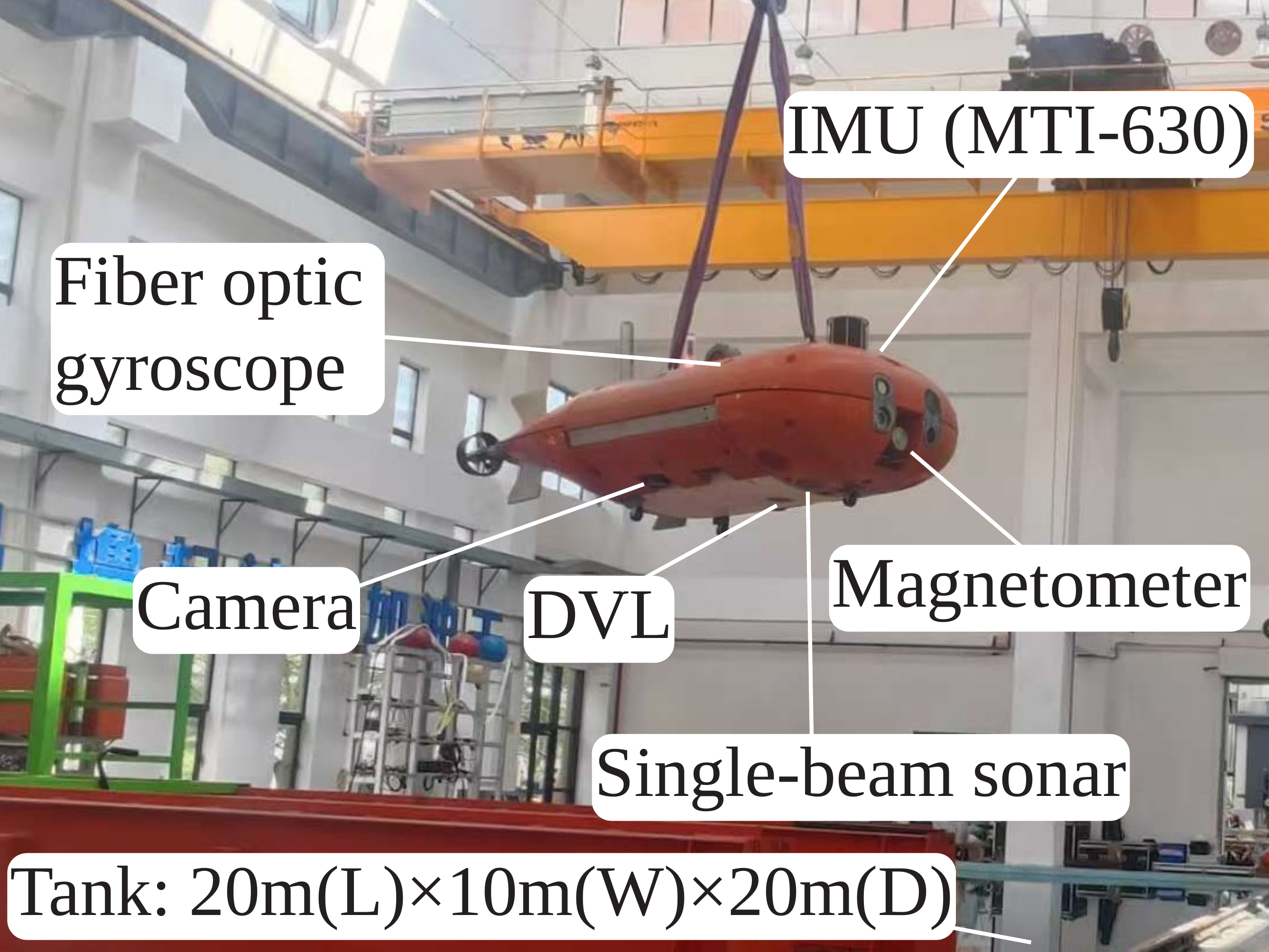}}    
\subfloat[Coil deployment ]{
    \label{figSensor}\includegraphics[width=0.45\linewidth]{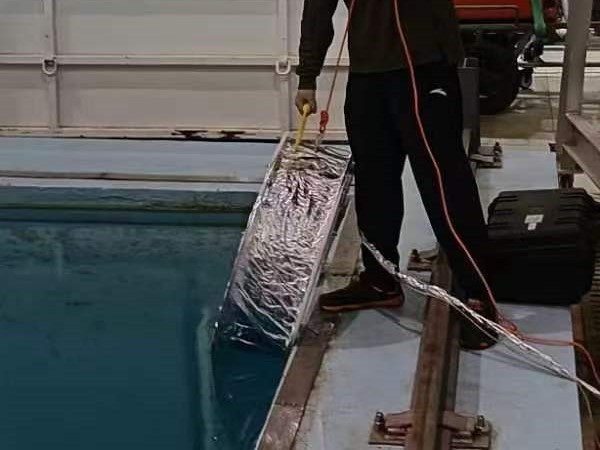}}
\caption{Experimental setup. We validate and evaluate the proposed VIMS on Hailong AUV in a water tank. Note that the equipped DVL is not used for state estimation in this study.}
\label{figTankSetup}
\end{figure}

%\begin{figure}[thpb]
%	\centering
%	\subfloat[Coil deployment]{
%		\label{figPlatform}\includegraphics[width=0.5\linewidth]{figs/07experiment/deploy_coil.jpg}}
%	\subfloat[Near-bottom AUV operation]{
%		\label{figSensor}\includegraphics[width=0.5\linewidth]{figs/07experiment/operation.jpg}}
%	\caption{The tank test}
%	\label{figTankTest}
%\end{figure}

\subsection{Loop Closure Verification}
\label{secLoop}
To verify loop closure, we match the 3D points tracked in the sliding window with 2D pixels in the visual submap candidates from visual place recognition. The process begins by replicating the 3D points across scales, as shown in Figure \ref{figORB}. Then, we encode them using ORB descriptors, discarding points with low Harris responses. The 2D-pixel points in the submaps have also been encoded during their construction (see Section \ref{secVisual}). Lastly, the 3D points are matched with 2D pixels in the submaps via feature correspondences using ORB descriptors. We apply a 2-nearest neighbors (2NN) test instead of a simple threshold to ensure reliable matches.

To further mitigate incorrect matches, we design a two-step geometric outlier rejection scheme:
\begin{itemize}
    \item 3D-2D matching is performed using the Perspective-n-Point (PnP) and RANSAC~\cite{lepetit2009ep}, discarding loop candidates with insufficient matched points.
    \item Covisibility constraints are checked, and loop candidates violating these constraints are eliminated.
\end{itemize}
% 3D-2D matching is conducted using the PnP test with RANSAC~\cite{lepetit2009ep}, discarding candidates with insufficient matched points. Second, we eliminate loop candidates that violate covisibility constraints.
\par For reliable loop closure candidates, we estimate relative pose measurements from the 3D-2D matching and integrate a loop closure factor into the pose graph.
%(see (\ref{eqClosure}))

\section{Experiments}

We evaluate VIMS through real-world experiments. Section \ref{sectSetup} first outlines the experimental setup followed by other sections for validation and evaluations.

% Section \ref{sectSetup} outlines the experimental setup, Section \ref{secProtocol} details the evaluation protocol, and Section \ref{sectEval} presents the results and numerical analysis.

%\yh{add on what other subsections doing. brief and compact.}

\subsection{Experimental Setup}
\label{sectSetup}

\begin{figure}[t]
\centering
\includegraphics[width=0.8\linewidth]{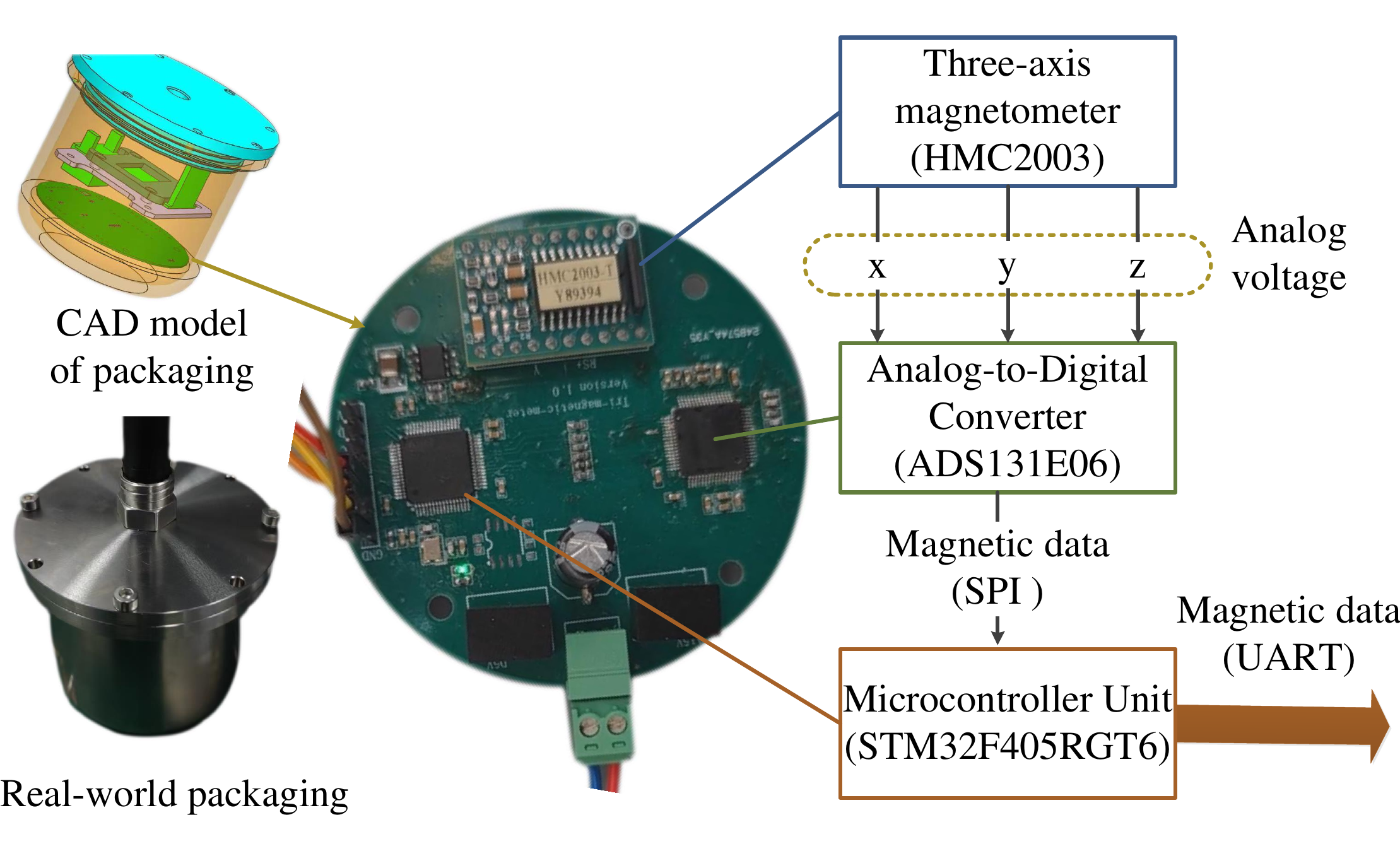}
\caption{Self-designed high-sampling-rate magnetometer. We use a 24-bit multi-channel ADC to measure magnetic fields, capturing data at a rate of 1000 Hz. The ADC communicates with a microcontroller unit (MCU) via the serial peripheral interface (SPI) protocol. After processing, the MCU transmits the data using the universal asynchronous receiver/transmitter (UART) protocol.}

%{An analog magnetometer (HMC2003) is employed for magnetic field measurement, integrated with a 24-bit multi-channel analog-to-digital converter (ADC) for accurate data acquisition at a sampling rate of \SI{1000}{\hertz}. The ADC communicates with the microcontroller unit (MCU) via the serial peripheral interface (SPI) protocol. Post data processing, the MCU transmits the acquired data using the universal asynchronous receiver/transmitter (UART) protocol. The entire system is housed in waterproof packaging, rated for depths up to \SI{1500}{\meter}. \yh{make it compact for paper shorten. Hardware design is not your key contribution}} 
\label{figMagnetometer}
\end{figure}

We conduct experiments in a tank using the Haihong II AUV to validate the proposed VIMS (Figure \ref{figTankSetup}). The sensor suite comprises a self-designed magnetometer (Figure \ref{figMagnetometer}), a monocular camera (DS-2ZMN0407), an IMU (MTI-630), and a single-beam sonar (VA 500). Additionally, we use other high-grade sensors, a fiber optic gyroscope (FOG, HsKINS-SG-4500C1) and a DVL (RDI-Pathfinder), to provide the ground truth for evaluation. In the experiments, the coil is \SI{0.2}{\meter} in height and \SI{0.6}{\meter} in diameter. It comprises 330 turns of wire, each with a cross-sectional area of \SI{2.0}{\square\milli\meter} and a resistance of \SI{6.6}{\ohm}. The experimental test lasts \SI{749.5}{\second}, covering a travel distance of \SI{131.4}{\meter}. The vehicle's distance during the test ranges from \SI{2.2}{\meter} to \SI{5.1}{\meter}, with a maximum speed of  \SI{0.38}{\meter\per\second}.

%We feed the coil with a current of \SI{2.9}{\ampere} at a frequency of \SI{11}{\hertz}.

%\begin{table}[htbp]
%	\renewcommand\arraystretch{1.2}
%	%   \setlength{\abovecaptionskip}{30pt}%    
%	\setlength{\belowcaptionskip}{0pt}%
%	\caption{Parameters of the coil}
%	\begin{center}
%		\begin{tabular}{ccccc}
    %			\toprule[1.5pt]
    %			Parameter &Coil size&Turns&Wire diameter&Resistor \\
    %			\midrule
    %			Value&\(\O\SI{0.85}{\meter}\times H \SI{0.60}{\meter}\)&260&\SI{1.5}{\milli\meter}&\SI{6.9}{\ohm}\\
    %			\bottomrule[1.5pt]
    %		\end{tabular}
%	\end{center}
%	\label{tableTestParam}
%\end{table}

\begin{figure}[t]
\centering
\subfloat[Image sequence]{
    \label{figSeq}\includegraphics[width=0.5\linewidth]{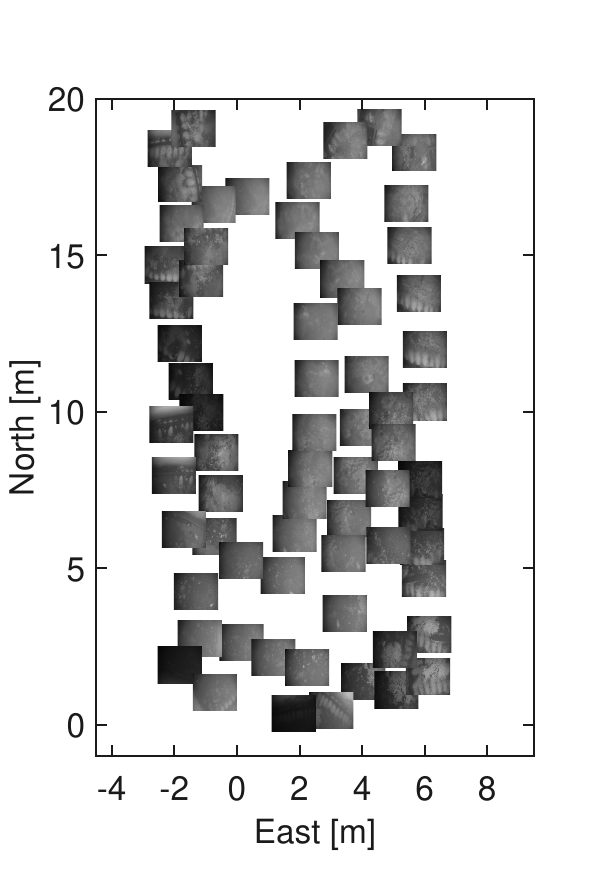}}
\subfloat[Mapping results ]{
    \label{figTankLandmark}\includegraphics[width=0.5\linewidth]{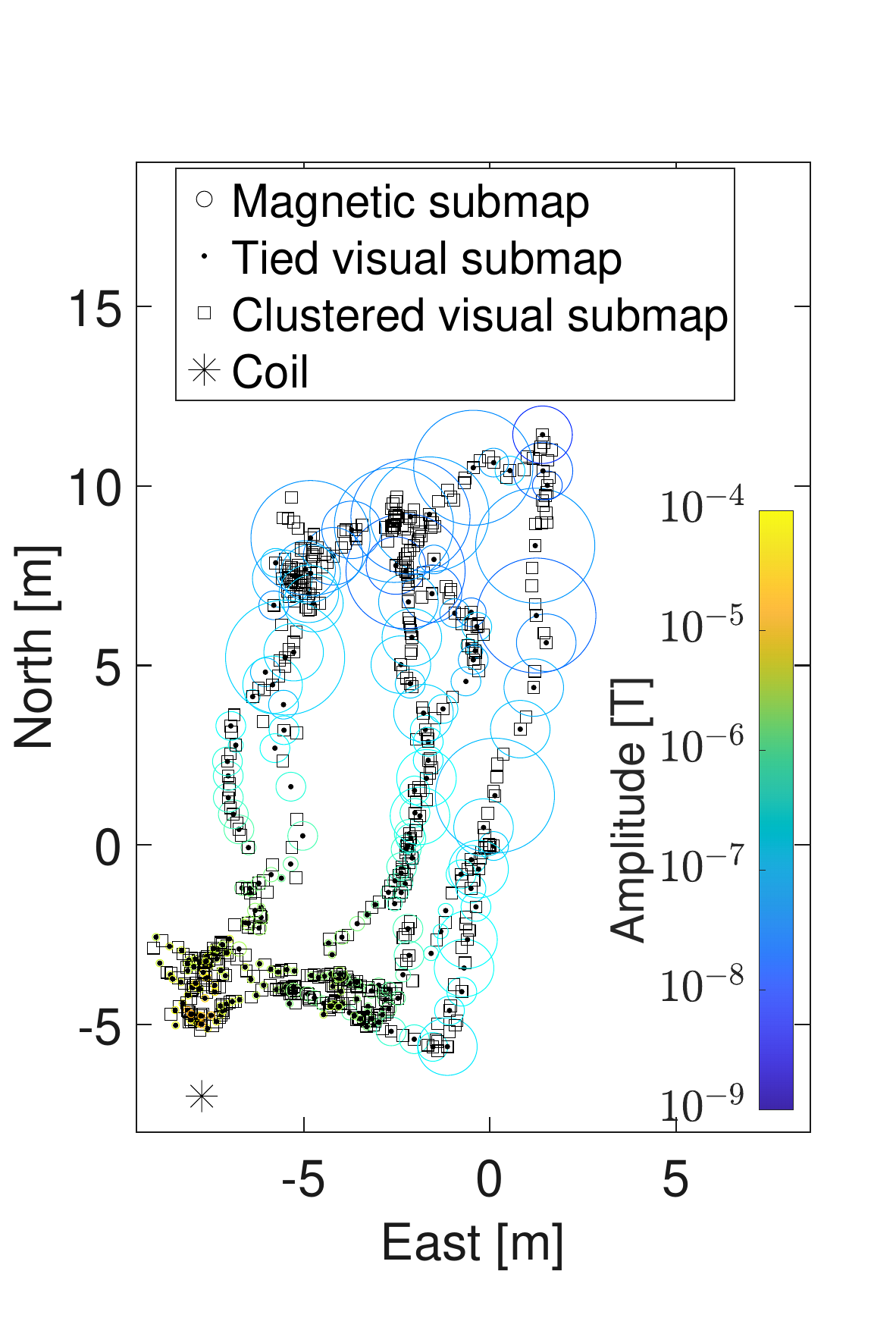}}		
\caption{Measured image sequence and mapping results. (a) Image sequence with respect to positions. The brightness of the images varies with the vehicle's distance to the bottom or tank wall. (b) Multi-scale map construction results in the tank test. The coil is positioned at the bottom of the tank, with its horizontal coordinates specified as $[-7.56, -7.20]^\top$ meters. Note that we assume the coil position is unknown for VIMS.}
\label{figMapping}
\end{figure}

%\begin{figure}[htpb]
%	\centering
%	\includegraphics[width=0.99\linewidth]{figs/07experiment/mvis_map3.pdf}
%	%	\subfloat[Multi-scale map]{
%		%		\label{figFieldLandmark}\includegraphics[width=0.99\linewidth]{figs/evaluation/field_landmark.pdf}}
%	%	
%	%	\subfloat[Magnetic amplitude map]{
%		%		\label{figFieldMagMap}\includegraphics[width=0.99\linewidth]{figs/08experiment/mag_map/field_mag_map.pdf}}
%	\caption{Multi-scale map construction results in the tank test. The coil is positioned at the bottom of the tank, with its horizontal coordinates specified as $[-7.56, -7.20]^\top$ meters. Note we assume the coil position is unknown for VIMS.}
%	\label{figTankLandmark}
%\end{figure}

%\begin{figure*}[htbp]
%	\includegraphics[width=0.99\linewidth]{figs/empty.png}
%	
%	\caption{Experimental results of different tests. \textbf{Left}: image sequence with respect to positions. \textbf{Middle}: trajectory estimates. \textbf{Right}: detected loops (the ``($\cdot$)/($\cdot$)'' notation represents the number of correct loops versus the total instances of loop verification).}
%	\label{figMultiResult}
%\end{figure*}

\begin{figure}[htpb]

\centering
\includegraphics[width=1.0\linewidth]{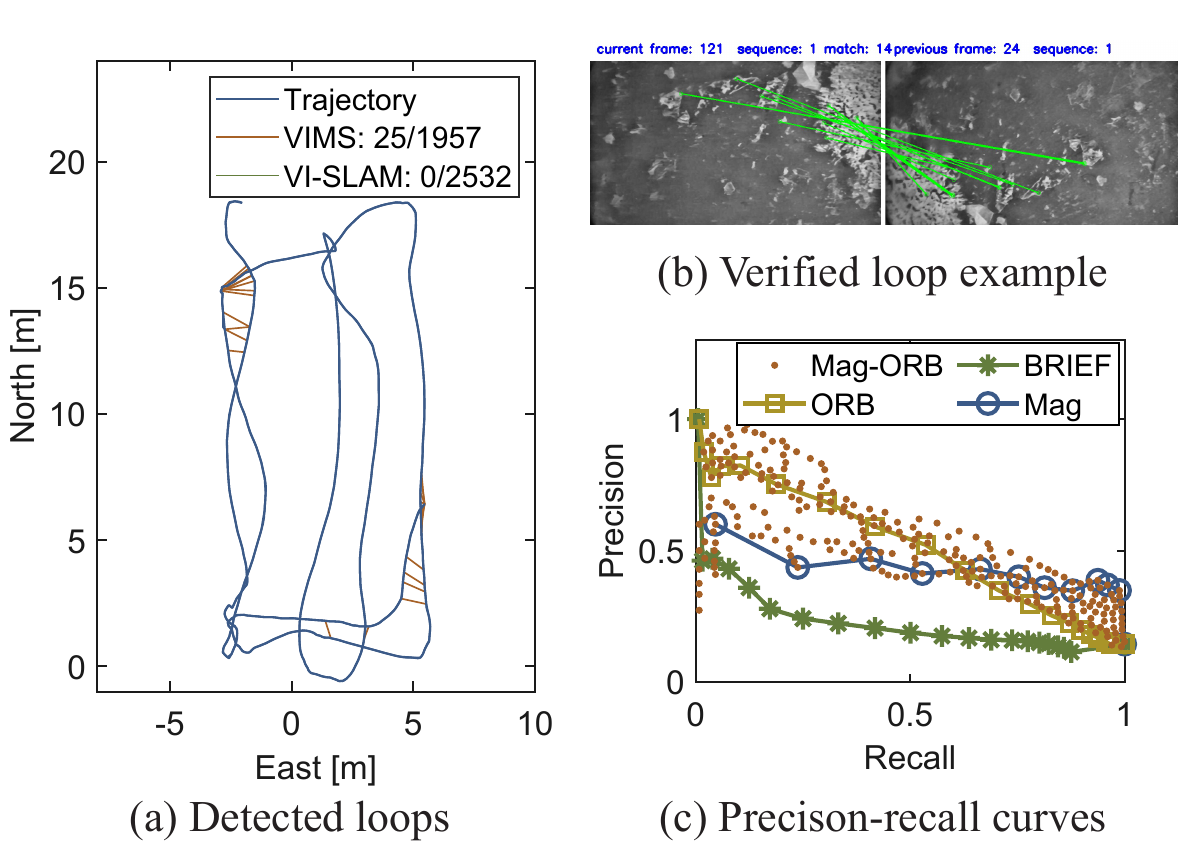}		
\caption{Loop closing results. (a) The notation ``($\cdot$)/($\cdot$)'' represents the number of correct loops versus the total instances of loop verification; (b) Illustration of a verified loop example obtained in VIMS; (c) Precison-recall curves of different place recognition methods, including Mag-ORB (magnetism \& ORB),  Mag (magnetism only), ORB, and BRIEF. Mag-ORB's adaptability forms precision-recall regions, unlike traditional curves. } 
\label{figLoopclosure}
\end{figure}

%\yh{Can you merge three trajectories into one figure? Better do it; if cannot, keep it as it is.}}
\begin{figure}[t]

\centering
\subfloat[{Mag-ORB}]{\label{figPrField}\includegraphics[width=0.5\linewidth]{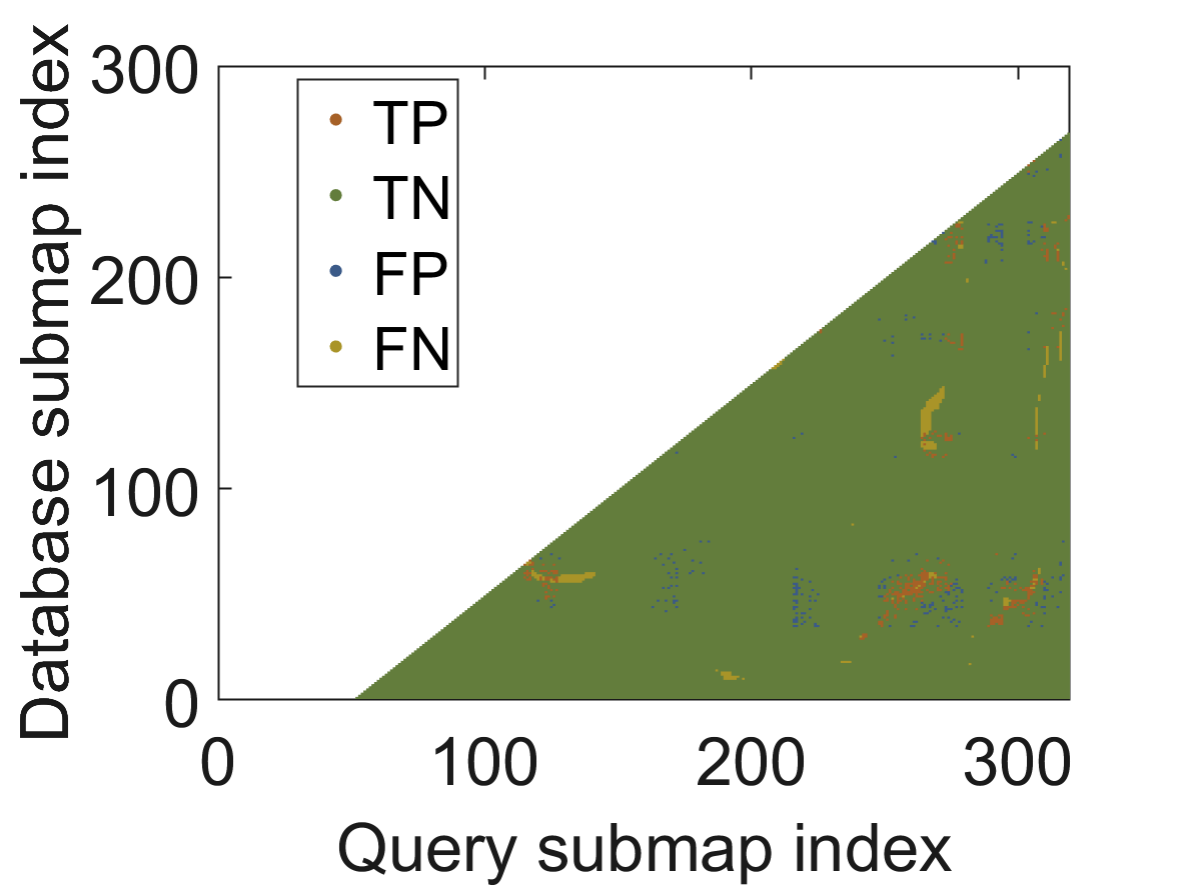}}
\subfloat[{Mag}]{
    \includegraphics[width=0.5\linewidth]{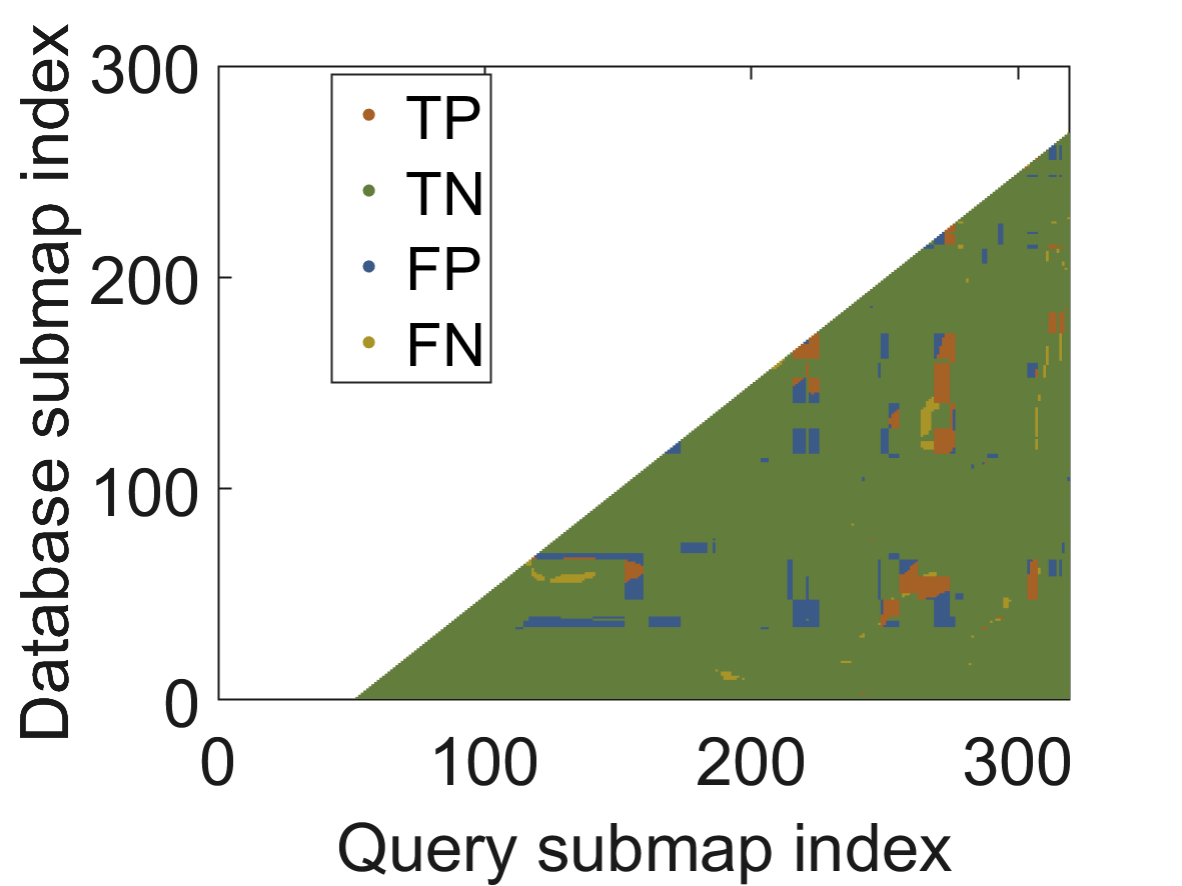}}
\\
\subfloat[{ORB} ]{
    \includegraphics[width=0.5\linewidth]{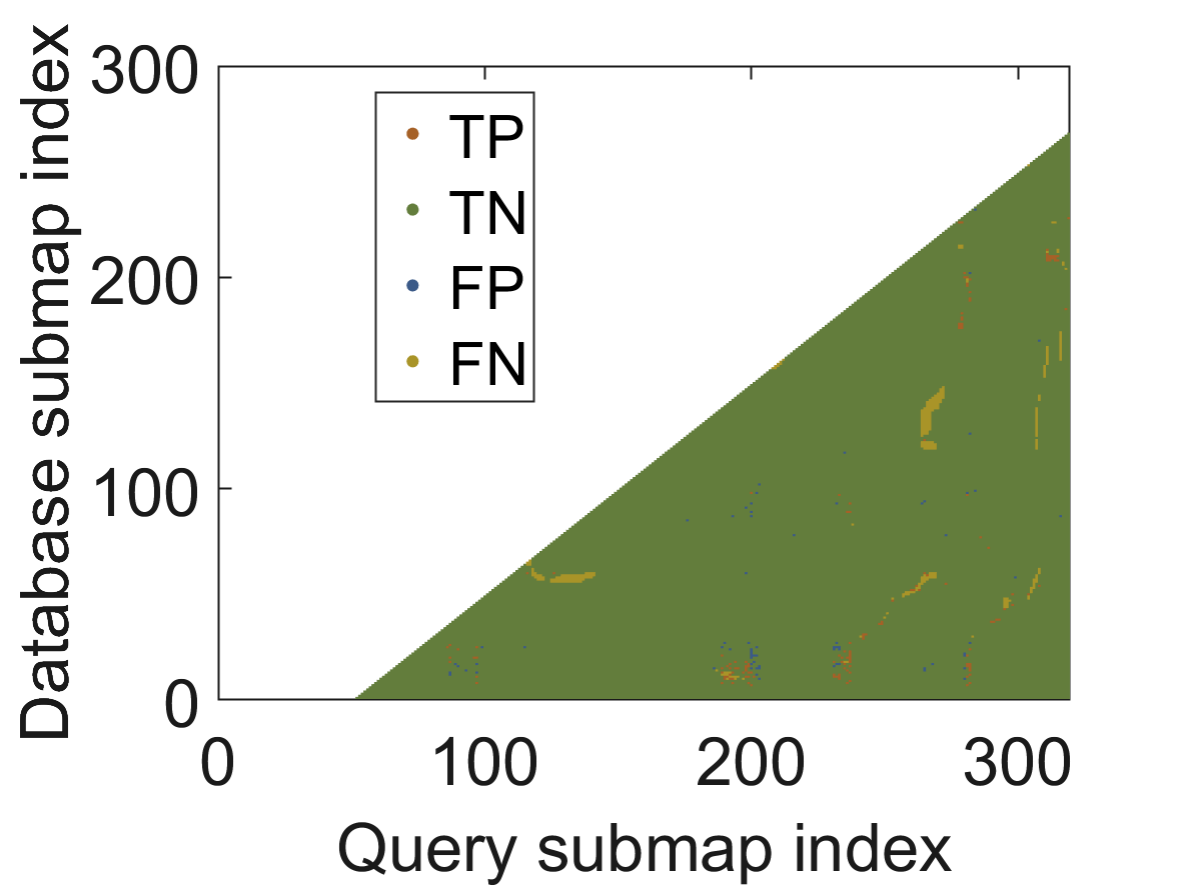}}
\subfloat[BRIEF ]{
    \includegraphics[width=0.5\linewidth]{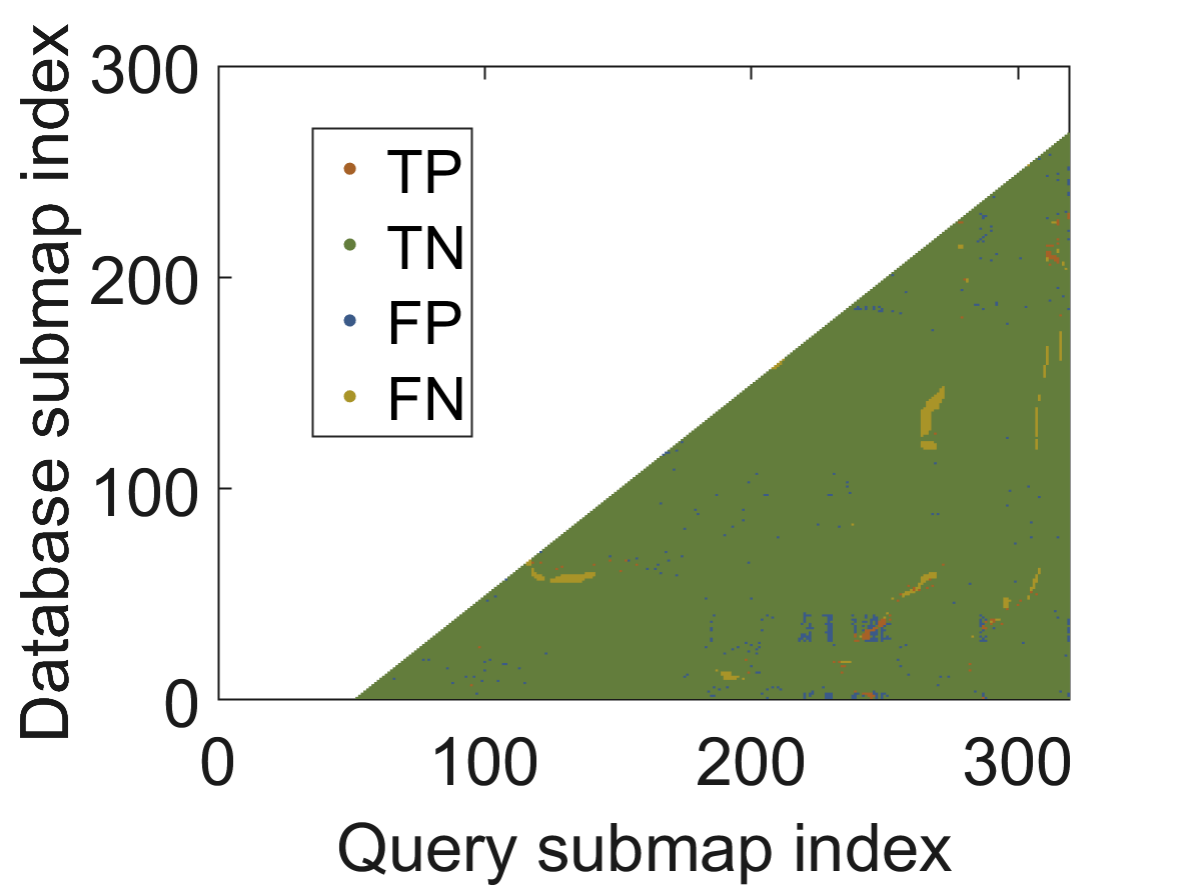}}		
\caption{Place recognition results using different methods. We utilize a \textit{recognition matrix} to visualize place recognition results, with the $x$-axis representing the most recent visual submap being queried, and the $y$-axis corresponding to the indices of historical submaps. Similarity thresholds for each method are set to maximize the $\text{F}_{0.5}$ score. Distinct recognition outcomes are depicted using different colors, excluding the region where $x < y + 50$ to mitigate the influence of consecutive submaps. The results show that Mag-ORB achieves the best precision-recall balance.
}
    \label{figMatrix}
\end{figure}
%\yh{please illustrate more on these figures. Very important. The reviewers do not want to guess.}
%\yh{this figure is not cited in the paper. also expand the explanation.}

\begin{figure}[t]

\centering

\subfloat[VIMS]{
    \includegraphics[width=0.45\linewidth]{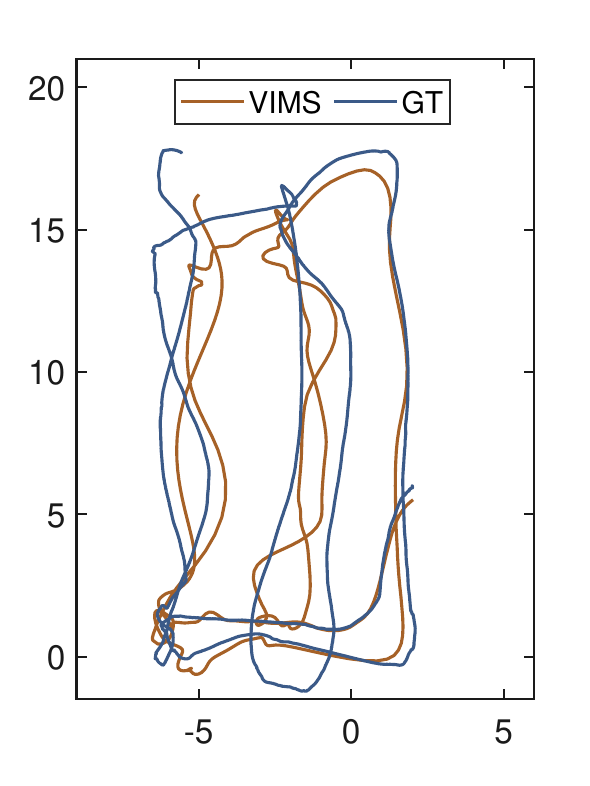}}
\subfloat[VI-SLAM]{
    \includegraphics[width=0.45\linewidth]{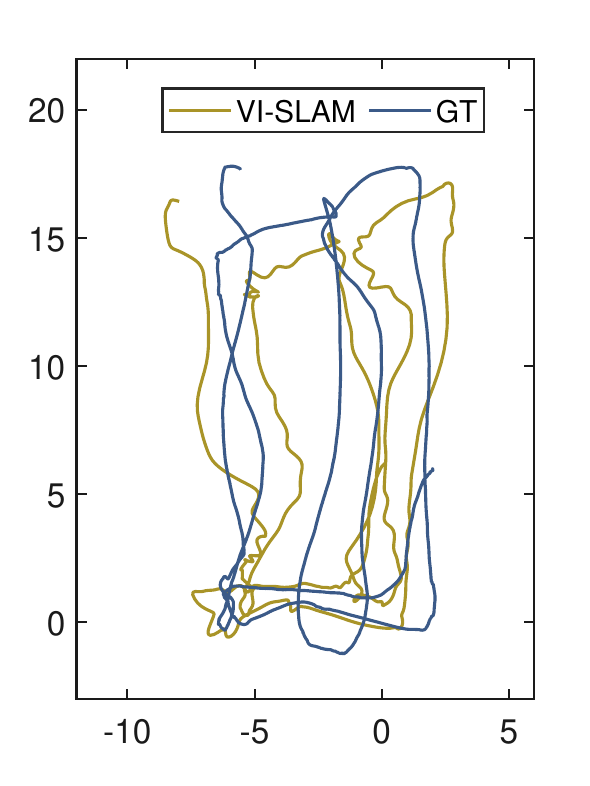}}
\caption{Trajectories of GT (Ground Truth), VI-SLAM and VIMS. }

    \label{figTraj}
\end{figure}

%The conducted experiments represent two types of underwater environements, natural and artificial environments. The image sequences with respect to the positions are shown in Figure \ref{figImgSeqField} and Figure \ref{figImgSeqTank}. Both of the two experimental scenerios are perceptually challenging. Natural underwater environments that contains such as soil, sand, stones, corals, etc. are fill with repetitive textures, . Artifical tank bottoms are typical feature-less areas, where the sparse textures of concretes are masked with soil and other sediments.   

\subsection{Evaluation Protocol}
\label{secProtocol}

To the best of our knowledge, there are currently no existing open-source localization or SLAM approaches that integrate both magnetism and vision. Classical visual-inertial methods, such as VINS-Mono~\cite{qin2018vins}, cannot operate effectively in challenging underwater environments. To build the baseline, we upgrade VINS-Mono in two ways with BRIEF descriptors~\cite{calonder2011brief}, and the enhanced VINS-Mono is referred to as \textit{VI-SLAM} hereafter. First, the original VINS-Mono algorithm verifies only the earliest visual place recognition candidate, which often leads to missed loops, particularly in underwater environments. To address this, we develop an enhanced loop closure strategy that verifies all candidates meeting a specified threshold, ensuring more robust loop detection for VI-SLAM. Second, the VIO algorithm struggles with feature depth estimation due to the low acceleration of the AUV. To mitigate this issue, we consistently integrate sonar priors into the local dead reckoning components of both the baseline VI-SLAM and our proposed VIMS approach, as detailed in Section \ref{sectVio}.

The performance is evaluated based on two criteria: loop closure and localization accuracy. Loop closure performance is measured using precision and recall rates of place recognition, as well as the total number of verified loops. Localization accuracy is assessed by comparing the 3D trajectory estimates to the ground truth.

\subsection{Performance Evaluation}	
\label{sectEval}
\subsubsection{Loop Closure Performance}

Figure \ref{figLoopclosure} and Figure \ref{figMatrix} present the difficulties faced by VI-SLAM. These challenges arise primarily because discrepancies in orientation significantly degrade the performance of BRIEF-based loop closing when revisiting nearby locations. The magnetism-based method demonstrates improved recall performance when a lower level of precision is tolerated. Overall, the integration of magnetism and vision achieves superior performance. As shown in Figure \ref{figLoopclosure}, VI-SLAM fails to detect any verifiable loops, whereas VIMS leverages hierarchical place recognition and ORB descriptors to enhance loop closing performance.

\subsubsection{Localization Accuracy}

Trajectories from VIMS and VI-SLAM are displayed in Figure \ref{figTraj}, aligned using evo tools~\cite{grupp2017evo}. The position RMSE values for VIMS and VI-SLAM are 1.11 meters and 1.56 meters, respectively, highlighting VIMS's better localization accuracy in underwater environments. Additionally, interframe matching in VI-SLAM struggles to correct yaw errors in feature-sparse areas, causing the estimated motion direction to deviate from the ground truth. In contrast, VIMS integrates geomagnetic measurements to improve yaw error correction and leverages enhanced loop closure to reduce pose errors.

\subsection{Mapping Results}

The multi-scale mapping results of the tank test are presented in Figure \ref{figTankLandmark}. The size of each circle represents the radii of submaps, as determined in Section \ref{secMagMap}. Notably, the magnetic submaps are depicted as 3D vectors, not merely by their total amplitude. This approach allows for overlapping among different magnetic submaps, enhancing their redundancy. The mapping results indicate that the proposed VIMS can provide visual and magnetic maps without prior knowledge of coil locations. The built maps can be reused for vehicle navigation to support long-term AUV autonomy, such as underwater teach and repeat missions~\cite{furgale2010visual}, thus reducing the need to deploy a full SLAM system for underwater navigation.

%\bb{teach and repeat means using available maps, we can make auv operates for long time?
% }
%\yh{the mapping results in 7(b) is not clear to me. I suggest present more zoomed in views.}
\subsection{Ablation Study}

To evaluate the contribution of each component, we conduct an ablation study of VIMS. Specifically, we test four variants of VIMS, each with a specific component disabled: the integration of sonar data, the fusion of alternating fields, the replacement of the BRIEF descriptor with ORB, and the incorporation of the geomagnetic field. The quantitative results are presented in Table \ref{tableAblation}.

\begin{table}[htbp]
	\renewcommand\arraystretch{1.2}
	\setlength{\belowcaptionskip}{0pt}%
	\caption{Quantitative Results of Ablation Study}
	\begin{center}
	\begin{tabular}{cccc}
            \toprule[1.5pt]
            Method &Loop closing&Trans. RMSE [m]&Rot. RMSE [deg]\\
            \midrule
            VI-SLAM &0/2532&1.56&9.70\\
            w/o Sonar& \XSolidBrush & \XSolidBrush & \XSolidBrush  \\
            w/o Alter.&14/2793&1.36&7.90\\
            w/o ORB &0/2557&1.37&8.99\\
            w/o Geom. &24/2150&1.33&8.00\\
            Full VIMS&25/1957&1.11&6.90\\
            \bottomrule[1.5pt]
        \end{tabular}
	\end{center}
	\label{tableAblation}
    \begin{tablenotes}
        \item  In the ``Loop closing'' column, the notation`` $(\cdot) / (\cdot)$'' represents the number of correct loops versus the total instances of loop verification. VI-SLAM can be regarded as equivalent to the combination of w/o Alter., w/o ORB, and w/o Geom, but with Sonar. \XSolidBrush indicates a failure of the system.
    \end{tablenotes}
\end{table}

The full VIMS, without any disabled enhancement components, achieves the best performance, indicating that each component contributes positively to the system. We can draw the following conclusions based on the ablation study:
\begin{itemize}

 \item \textbf{Integration of sonar data.} The absence of sonar data causes VIO-based dead reckoning to fail. This failure is attributed to the limited acceleration of the AUV (see Figure~\ref{fig_teaser}), which prevents accurate motion scale estimation, as noise and bias overwhelm the self-acceleration information in the acceleration measurements. 
    
    \item \textbf{Fusion of alternating field.} Disabling the fusion of the alternating field causes a drop in performance for both loop closing and pose estimation. This highlights the importance of the alternating field in achieving accurate localization and navigation.
    
   \item \textbf{Replacement of BRIEF with ORB.} Replacing BRIEF with ORB enhances loop closing performance, particularly in scenarios involving significant orientation changes when revisiting certain locations. Such notable orientation changes can be observed almost every time previously visited places are encountered.

    \item \textbf{Incorporation of geomagnetic field.} When the geomagnetic field is not incorporated, drift errors in orientation estimation increase, leading to a marked degradation in localization performance. This underscores the critical importance of integrating the geomagnetic field into the system. 

\end{itemize}

\section{Conclusion}
\label{secConclusion}

This study presents a novel and full SLAM system that integrates visual, inertial, magnetic, and sonar measurements, enabling effective vehicular localization in perceptually degraded underwater environments. Real-world experiments demonstrate its superiority over the baseline, specifically enhanced visual-inertial SLAM systems. Compared to our previous work~\cite{10778419}, the advancements introduced here enable SLAM to handle more complex and higher-speed AUV maneuvers, marking a significant step toward future field deployment. Future studies will focus on developing a more refined system and conducting in-depth variance estimation of motion to further improve state estimation performance. Additionally, we plan to open-source the experimental data to foster underwater SLAM research within the community.
%In the future, more detailed ablation studies will be conducted.
% In our current SLAM system, while errors are uniformly distributed across motions between consecutive frames post-loop detection, the magnitude of these errors fluctuates based on the visual quality of the data. Going forward, we aim to develop a more refined model that more accurately accounts for variations in visual input quality in our error estimation process.

% \bb{Maintaining the page limit is really challenging, and we still need to add author part.}

\bibliographystyle{IEEEtran}
\bibliography{IEEEabrv,IEEEexample}
% \bibliography{IEEEexample}

\end{document}